\documentclass[10pt,journal,compsoc]{IEEEtran}

\ifCLASSOPTIONcompsoc
  \usepackage[nocompress]{cite}
\else
  \usepackage{cite}
\fi

\ifCLASSINFOpdf
\else
\fi

\usepackage{url}

\usepackage{graphicx}
\usepackage{comment}
\usepackage{amsmath,amssymb} 
\usepackage{color}
\usepackage{lipsum}
\usepackage{mathtools, cuted}

\usepackage{algorithm}
\usepackage{algcompatible}

\usepackage[utf8]{inputenc}
\usepackage[english]{babel}
\usepackage{adjustbox}
\usepackage{import}
\usepackage[noend]{algpseudocode}
\usepackage{booktabs}
\usepackage{relsize}
\usepackage{subcaption}
\usepackage{bm}
\usepackage{mathtools}
\usepackage{multirow}
\usepackage{graphicx}
\usepackage{array}
\usepackage[low-sup]{subdepth}
\usepackage{xspace}
\usepackage{bm}
\usepackage{babel}
\usepackage[font=small,labelfont=bf]{caption}
\usepackage{booktabs}
\usepackage{varwidth}
\newsavebox\tmpbox

\allowdisplaybreaks

\DeclareCaptionFont{blah}{\small\fontseries{n}\fontfamily{phv}\selectfont}
\DeclareCaptionLabelSeparator*{twonl}{\\[0.1\baselineskip]}
\makeatother

\newcommand{\floordiv}{\mathbin{\!/\mkern-5mu/\!}}
\newcommand{\ours}{PAMELA\xspace}

\begin{document}

\title{Meta-learning the Learning Trends Shared Across Tasks}

\author{Jathushan Rajasegaran, Salman Khan, Munawar Hayat, Fahad Shahbaz Khan, Mubarak Shah
\IEEEcompsocitemizethanks{\IEEEcompsocthanksitem Jathushan Rajasegaran, Munawar Hayat, Salman Khan and Fahad Shahbaz Khan are with Inception Institute of Artificial Intelligence, UAE.\protect\\
E-mail: firstname.lastname@inceptioniai.org
\IEEEcompsocthanksitem Mubarak Shah is with Center for Research in Computer Vision, University of Central Florida, US.}
}

\markboth{Preprint}%
{Shell \MakeLowercase{\textit{et al.}}: Bare Demo of IEEEtran.cls for Computer Society Journals}

\IEEEtitleabstractindextext{%
\begin{abstract}
Meta-learning stands for {`learning to learn'} such that generalization to new tasks is achieved. Among these methods, Gradient-based meta-learning algorithms are a specific sub-class that excel at quick adaptation to new tasks with limited data. This demonstrates their ability to acquire transferable knowledge, a capability that is central to human learning. However, the existing meta-learning approaches only depend on the current task information during the adaptation, and do not share the meta-knowledge of how a similar task has been adapted before. To address this gap, we propose a \emph{`Path-aware’} model-agnostic meta-learning approach. Specifically, our approach not only learns a good initialization (meta-parameters) for adaptation, it also learns an optimal way to adapt these parameters to a set of task-specific parameters, with learnable update directions, learning rates and, most importantly,
the way updates evolve over different time-steps. Compared to the existing meta-learning methods, our approach offers the following benefits: (a) The ability to learn gradient-preconditioning at different time-steps of the inner-loop, thereby modeling the dynamic learning behavior shared across tasks, and (b) The capability of aggregating the learning context through the provision of direct gradient-skip connections from the old time-steps, thus avoiding overfitting and improving generalization.  In essence, our approach not only learns a transferable initialization, but also models the optimal update directions, learning rates, and task-specific learning trends. Specifically, in terms of learning trends, our approach determines the way update directions shape up as the task-specific learning progresses and how the previous update history helps in the current update. Our approach is simple to implement and demonstrates faster convergence compared to the competing methods. We report significant performance improvements on a number of datasets for few-shot learning on classification and regression tasks. Our codes are available at: {\url{https://github.com/brjathu/PAMELA}}.
\end{abstract}

}

\maketitle

\IEEEdisplaynontitleabstractindextext

\IEEEpeerreviewmaketitle

\IEEEraisesectionheading{\section{Introduction}\label{sec:introduction}}
Leveraging from their prior experience, humans can easily learn new concepts from a few observations. Few-shot learning aims to mimic this astounding capability, and requires quick model adaptation using only a few examples. In contrast, contemporary deep learning models are data hungry by nature,  learn each task in isolation and require significant training time. Meta-learning comes as a natural solution to this problem, due to its focus on `\emph{learning to learn}' a generalizable model from multiple related tasks. This strategy offers a quick adaptation to new tasks.

Model-agnostic meta-learning (MAML) \cite{finn2017model} is a popular approach that learns a generalizable representation which can be quickly adapted to a new task with only a few examples. In MAML, the meta-training operates in two nested loops. First, the task-specific parameters are learned in the inner-loop, followed by learning a shared set of parameters that acts as a good prior for all the tasks. Although MAML uses standard gradient descent for inner-loop optimization, recent works demonstrate that an improved inner-loop optimization can positively influence the performance \cite{li2017meta,park2019meta,antoniou2018train}. However, these existing methods do not consider the learning trends (update direction, step-size and evolution through iterations) in the inner-loop across different tasks.

In this paper, we develop a new \emph{model-agnostic} meta-learning framework called `Path-Aware MEta-LeArning’ (\ours). Compared to MAML, which learns a good initialization for the meta-learner, our approach learns the optimal learning trend shared across different tasks. This means that, in addition to a transferable initialization, we learn the update directions, the learning rates and, most importantly, the way updates evolve over different time-steps. As an example, our model can encode how the inner-loop training first commences with large steps and converges to shorter steps as it gets close to the local minima. Similarly, it can also learn how to optimally reuse task parameters from the past updates. In essence, our approach provides a new way to encode the prior (generalizable) knowledge in a more principled and flexible manner.

\ours showcases two main novelties to meta-learn the learning paths shared across tasks. \emph{First,} we propose to learn distinct gradient preconditioning matrices at different iteration steps in the inner-loop. Pre-conditioning was first  proposed in Meta-SGD \cite{li2017meta}. However, as opposed to \cite{li2017meta}, which learns a single preconditioning matrix for the whole path, we show that learning \emph{iteration-specific} preconditioning provides us the flexibility to model varying trends along the learning paths. \emph{Second,} to learn better context at each time step, we propose a residual connection based gradient preconditioning that allows multiple old gradients to directly flow via the gradient-skip connections. This approach not only provides better context, but also helps avoid gradient vanishing (and model overfitting) for the task at hand.

Overall, our approach helps in providing an improved modelling of the short-term knowledge specific to a task and also the long-term trends shared between tasks. In a similar pursuit, recurrent neural network (RNN) based meta-learning methods have been explored in the literature \cite{hochreiter2001learning,andrychowicz2016learning,ravi2016optimization}. Andrychowicz  \emph{et al.}~\cite{andrychowicz2016learning} proposed an LSTM meta-learner that learns to mimic a gradient based optimizer, outputting updates at different time steps. Ravi and Larochelle \cite{ravi2016optimization} extended the LSTM meta-learner for few-shot settings, where both the base initialization and update mechanism are learned, resulting in a high complexity. In contrast to these approaches, we take a different perspective on aggregating path context. We learn unique trends in each update step while simultaneously combining past context using gradient-skip connections.  This results in an easy-to-train model with faster convergence and superior performance.

\section{Related Work}
Meta-learning algorithms can be grouped into three main categories: \textbf{a)} Metric-based, \textbf{b)} Model-based and \textbf{c)} Gradient-based optimization methods. We outline these below.

\vspace{0.1cm}
\noindent \textbf{Metric and Model-based Meta-learning:} Metric-based methods learn an optimal distance kernel, parameterized by meta-parameters. Koch~\emph{et al.}~\cite{koch2015siamese} used a siamese network to compare the distance between samples. Vinyals \emph{et al.}~\cite{vinyals2016matching} proposed matching networks, which learn the similarity between support sets and the test samples by producing a weighted sum of the support set labels and attention kernel. Relation networks~\cite{sung2018learning} use a learnable module to predict relation scores in the feature space. Further, prototypical networks~\cite{snell2017prototypical} define a prototype for each task as an average embedding of the network output, and learn the metric space via error back-propagation. The model-based meta-learning algorithms learn to quickly adapt to the tasks by either having an internal memory or by using fast weights. Santoro \emph{et al.}~\cite{santoro2016meta} proposed Memory-Augmented Neural Networks which use a Neural Turing Machine to store new information, thus enabling adaptation to new tasks with few samples. Meta Network~\cite{munkhdalai2017meta} uses a combination of fast and slow weights to rapidly adapt to new tasks.
\vspace{0.1cm}
\noindent \textbf{Gradient-based Meta-learning:} Meta-LSTM~\cite{ravi2016optimization} models the meta-learner as a recurrent network (LSTM), thereby learning to generate the parameters of the learner at each time step. Finn \emph{et al.}~\cite{finn2017model} proposed a Model Agnostic Meta-Learning (MAML) algorithm. MAML functions in two loops (inner and outer) and  uses the second-order gradients to update the initial parameters. This requires backpropagation through the learning steps of the inner loop. Therefore, to reduce the computational complexity of MAML, the authors also proposed a first-order approximation of MAML (FOMAML), which considers only the last gradient on the inner loop update as the meta-gradient. In addition to FOMAML, Nichol~\emph{et al.}~\cite{nichol2018first} used Reptile, which is a better first-order approximation since it computes an average of all inner loop gradients for the meta-update. Reptile is more robust to sample selection and gradient noise. In addition to this,~Antoniou~\emph{et al.}~\cite{antoniou2018train} made few changes to MAML, such as layer-wise learning rates and accumulating meta-loss in each update. Although modeling the inner-loop updates with layer-wise learning rates is similar to our work, we have more flexible updates with kernel-wise learning rates and our gradient-skip connections help to minimize the vanishing meta-gradients. Rajeswaran~\emph{et al.}~\cite{rajeswaran2019meta} used a closed-form approximation to solve the MAML optimization, using a regularized loss function. Khodadadeh~\emph{et al.}~\cite{khodadadeh2019unsupervised} proposed a semi-supervised version of MAML algorithm. However, these MAML based methods treat the inner loop as a fixed learning process and do not share any meta knowledge during the learning updates.

Meta-SGD~\cite{li2017meta} preconditions the inner loop gradients with meta-trainable learning rate, thus allowing the meta-information to be used in the inner loop adaptation process. Similar to Meta-SGD, Park~\emph{et al.}~\cite{park2019meta} proposed Meta Curvature, which learns the curvature information to precondition the gradients. Meta-Curvature performs better than Meta-SGD in practice, however, it requires higher-order tensor operations to learn the curvature. Recently, Flennerhag~\emph{et al.}~\cite{flennerhag2019meta} proposed a warp gradient-based approach, which inserts warp layers in between the network layers, and updates these after a specific interval using the accumulated gradients (meta-update). However, WarpGrad~\cite{flennerhag2019meta} is not model agnostic since it requires modifications in the model architecture. Further, WarpGard preconditions the feature space in the forward pass, while our method preconditions only the gradients, thereby making it totally independent from the model architecture. In addition, different to the above mentioned approaches, \ours \ learns the optimal inner-loop update trends shared across tasks as well as the best combination of knowledge acquired at previous steps.

\begin{figure*}[t]
    \centering
    \includegraphics[angle=-90, scale=0.30]{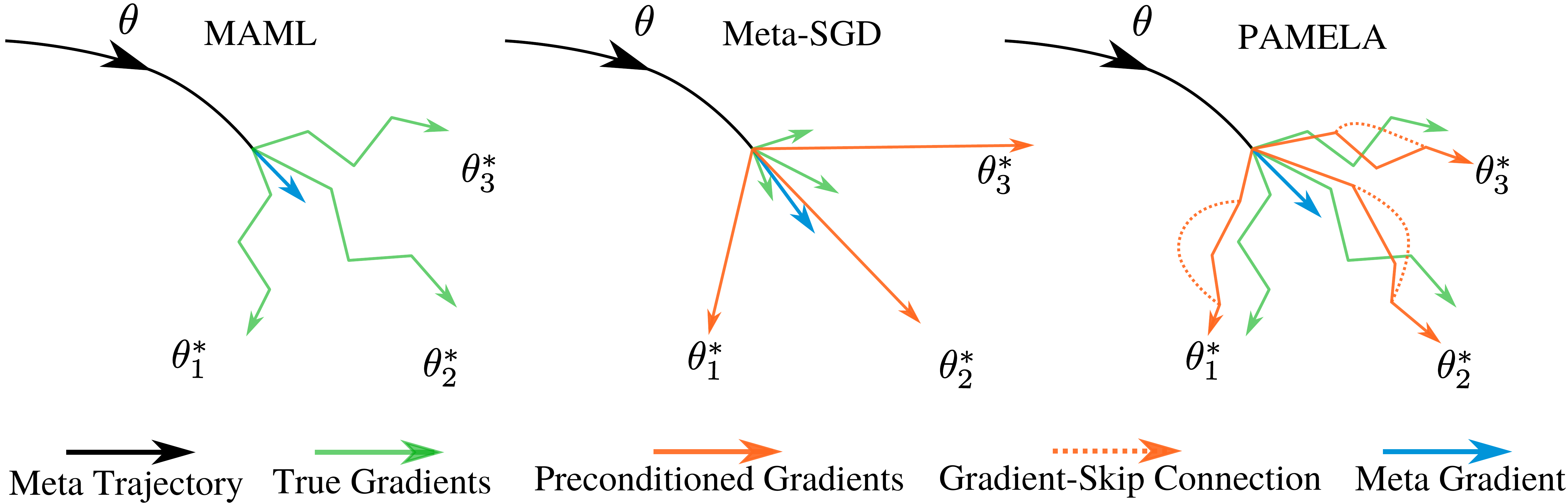}
    \caption{\emph{Inner-loop optimization of model-agnostic gradient-based meta-learning algorithms:} MAML takes multiple steps towards the optimal parameters $\theta^*$, and Meta-SGD takes a single step with a preconditioned direction. While \ours\ takes multiple steps towards $\theta^*$ with a unique preconditioned direction for each step, as well as shares the meta information between past and future updates via Gradient-skip connections. This results in a faster convergence closer to the optimal parameters.}
    \label{fig:info}
\end{figure*}

\section{Path-aware Meta-learning}

In a classical learning setting, a model learns knowledge about the training set and applies it on the test set. This paradigm learns each task in isolation and demands large quantities of data and training time for each training cycle. In contrast, meta-learning seeks to learn about learning so that quick adaptation to new tasks is possible. In this pursuit, we propose a new model-agnostic meta-learner called \ours, which not only learns a better initialization that can generalize across tasks, but also models the learning trends that reveal how the inner-loop (task-specific update process) evolves during training.
Further, it uses the acquired meta-knowledge to combine previous gradients for a stronger context, thereby converging to a better initialization.

\subsection{Meta-training}
Consider a classification model $f_{\theta}$, which is parameterized by $\theta$, and $\mathbf{F}_{\Phi}(\theta)$, which is an inner-loop optimization function of the meta-learner parameterized by $\Phi$. We suppose that a task $\mathcal{T}$ contains two sets of data, a training set $\mathcal{D}_{tr}$ and a validation set $\mathcal{D}_{val}$ i.e., \{$\mathcal{D}_{tr}, \mathcal{D}_{val}\}\in \mathcal{T}$. We want to optimize $f_{\theta}$ on each task $\mathcal{T}\sim {P}(\mathcal{T})$, so that after learning from multiple tasks of similar nature, $\mathbf{F}_{\Phi}$ can figure out how to rapidly learn a new task sampled from the same distribution ${P}(\mathcal{T})$. This optimization problem is given by:
\begin{align}
\underbrace{
    \min_{\theta, \Phi} \;\mathbb{E}_{\mathcal{T}\sim {P}(\mathcal{T})}\big[\mathcal{L}_{\mathcal{D}_{val}}( f_{\underbrace{\mathbf{F}_{\Phi}(\theta)}_{\text{inner-loop}} })\big]}_{\text{ outer-loop}},
    \label{eq:main}
\end{align}
where $\mathcal{L}$ is a loss function for the given task.   As Eq.~\ref{eq:main} shows, meta-training consists of two parts: an inner-loop and an outer-loop. The inner-loop learns the task-specific adaptation, while the outer-loop learns about the `learning process of the inner-loop'. Below, we respectively describe the two nested loops.

\subsubsection{Inner-loop Optimization}  In \ours, the inner-loop optimization function $\mathbf{F}_{\Phi}$ is defined as an iterative update process. $\mathbf{F_{\Phi}}$ takes the model parameters $\theta$ and gives $\theta_n$ after $n$ inner-loop iterations, such that $\mathcal{L}_{\mathcal{D}_{tr}}(f_{\theta})$ is minimized,
\begin{align}
    \theta_n \leftarrow \mathbf{F}_{\Phi}(\theta) = \min_{\theta}\; \mathbb{E}_{\mathcal{D}_{tr}\sim \mathcal{T}}[\mathcal{L}_{\mathcal{D}_{tr}}(f_{\theta})].
\end{align}

 The inner-loop learning of \ours\ differentiates it from other gradient-based meta-learning algorithms such as  MAML~\cite{finn2017model} and Meta-SGD~\cite{li2017meta}. MAML~\cite{finn2017model} simply follows the gradient descent during inner-loop optimization, hence $\theta$ moves along the true gradients.
 The online update is formally given by: $\theta_{j+1} = \theta_{j} - \alpha \nabla_{\theta_{j}} \mathcal{L}_{\mathcal{D}_{tr}}(f_{\theta_{j}})$, where $\alpha$ is a constant scalar and $j>0$.  However, following the true gradients on a limited number of samples will not converge to a globally optimal set of parameters, which minimizes the loss on all $\mathcal{D}_{tr} \in \mathcal{T}$. In summary, MAML~\cite{finn2017model} uses no gradient pre-conditioning, thus ignoring \emph{how other tasks update} their parameters.  To improve MAML, Meta-SGD preconditions the true gradients by considering $\alpha$ (step size) as a meta-parameter. As a result, $\theta$ moves in a direction that can provide the optimal parameters for all tasks.  Although Meta-SGD moves $\theta$ along a  meta-learned direction, it takes only a single step towards the optimal parameters and therefore does not guarantee convergence. Notably, Meta-SGD cannot be simply run with multiple steps since it uses a single preconditioning matrix that cannot model the different directions required to be learned for multiple-steps in the inner-loop. As an example, early updates generally need stronger gradients, while later ones do not; thus, a single parameter cannot encompass the step-wise learning behavior. To validate this, we extend the Meta-SGD with multiple inner-loop steps and notice that it fails to converge with more iterations.

In contrast to the above methods, \ours\ preconditions the true gradients in order to learn \emph{unique} step-wise update directions and to \emph{share} meta-knowledge between different steps. This behavior is illustrated in  Fig.~\ref{fig:info}.
Consequently, our inner-loop optimization function $\mathbf{F}_{\Phi}$ has two sets of learnable meta-parameters, $\bm{Q}$ and $\bm{P}^w$, respectively:
\begin{align*}
 \Phi &= \{\bm{Q}, \bm{P}^w \} , \, s.t.,\;  \bm{Q} =  \{Q_0,Q_1, ... Q_{n-1}\}  \text{ and } \\ \bm{P}^w &= \{P^w_0, P^w_w, ... , P^w_{(n \, \floordiv\, w - 1)*w}\},
\end{align*}
where $\floordiv$ denotes floor division, $n$ is the total number of inner-loop updates and $w$ is the interval size over which we  aggregate context using gradient-skip connections. Each $Q$ and  $P$ is a  vector, with dimensions same as the corresponding layer parameters it represents.
The parameters $Q_j$ learn the trends {for} \emph{each} inner-loop update, while $P^w_j$ learn the correlations \emph{between} current and past updates.

\begin{figure*}[t!]
    \centering
    \includegraphics[angle=-90, scale=0.33]{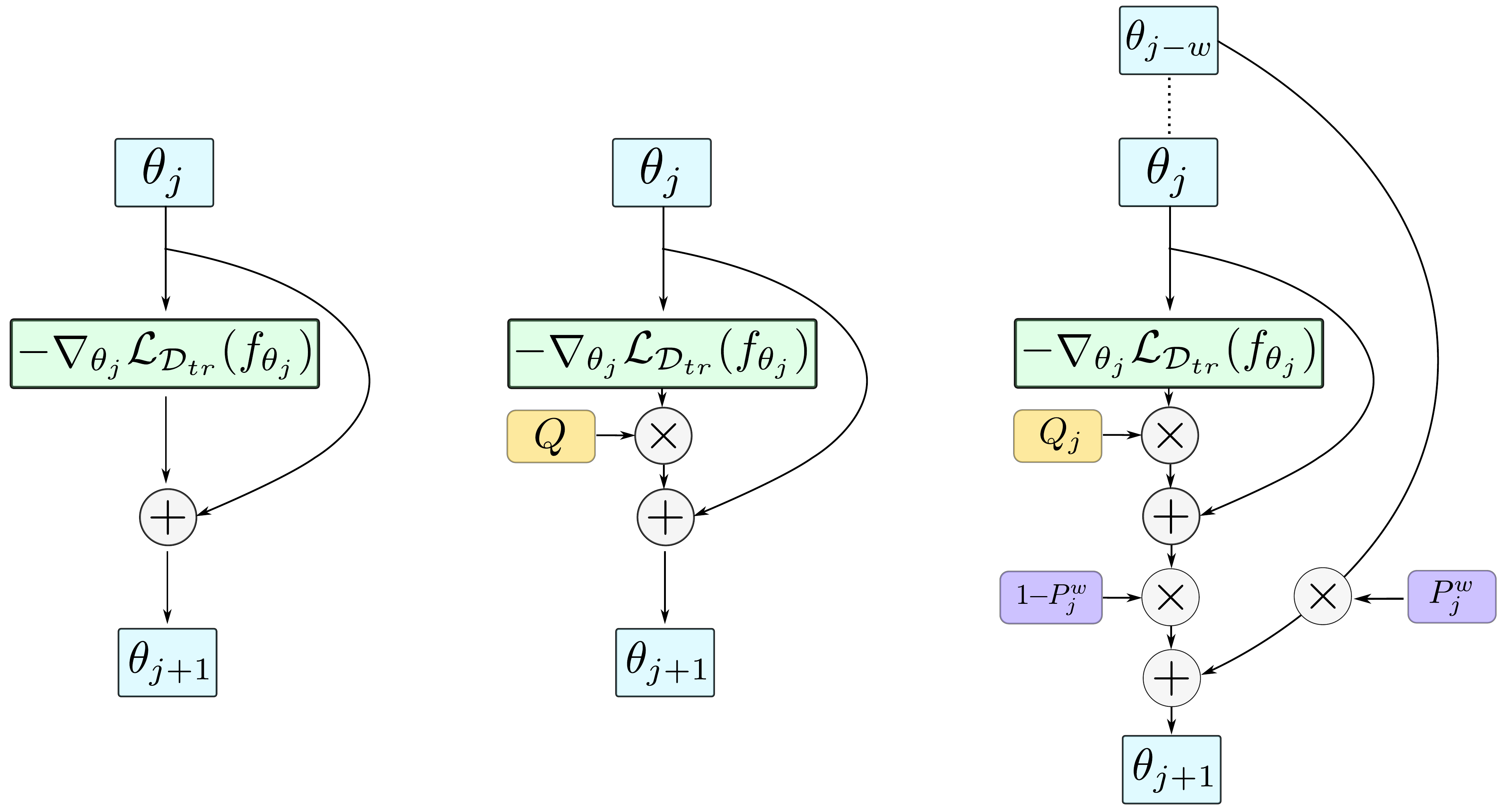}
    \caption{\emph{Comparison between MAML, MetaSGD and  our \ours}. \emph{Left:} One step inner-loop update of MAML, which can be understood as an identity gradient-skip connection of the model parameters and learning the gradients as residual signals. However, the gradients are not pre-conditioned; therefore, the whole adaptation process is independent of the meta parameters. \emph{Middle:} Meta-SGD uses a single matrix $Q$ to pre-condition inner-loop gradients. However, this will lead to faster jumps at the end of multiple update steps and potentially collapse the model. \emph{Right:} \ours uses a separate pre-condition matrix $Q_j$ for each update step, as well as sharing the information from the previous learned parameters using a longer interval of gradient-skip connection with a coefficient $P^w_j$, thereby learning the update-trends shared across tasks. }
    \label{fig:comp}
\end{figure*}

\noindent \textbf{$\bm{Q}$: Modeling the Learning Directions:}
Each element $Q_j \in \bm{Q}$ controls the learning trend of $f_{\theta_j}$ in each inner-loop update step, thus discretely learning and sharing the meta-knowledge between tasks at different time steps.
At the $j^{th}$ inner-loop update, we pre-condition the true gradients $\nabla_{\theta_j}\mathcal{L}_{\mathcal{D}_{tr}}(f_{\theta_j})$ by $Q_j$ using the Hadamard product. We use the term true gradients to denote the gradients calculated from the loss function, without any preconditions or projections. Therefore, the model parameters move in a direction that jointly encompasses the true gradient directions across tasks. However, assigning a single meta-parameter for each parameter in the model will suffer from large memory requirements. For example, $n$ inner-loop updates would require $n$-times more memory. Due to this, we only learn a single meta-parameter for each convolutional kernel. This choice is made because a kernel captures a single feature in the given images, therefore, all the parameters for a given kernel should adapt similarly as the model evolves in the inner-loop. For fully-connected layer, we keep a single meta-parameter for each parameter during inner-loop updates.

\noindent \textbf{$\bm{P}^w$: Modeling the Update Context and Learning Rates:}
The meta-parameters $P^w_{j}$ learn the correspondence between the past and the current gradient updates. For example, $P^w_{j}$ will fuse the knowledge from $\theta_{j-w}$ to $\theta_{j}$. Here, $w$ (interval-size) is a  hyper-parameter that controls the gap between the current and historical gradient update used to calculate the context at a given iteration. Interval size $w$ is fixed through out the learning process and the optimal value is set using a validation set. Finally, we use the following update rule for the $j^{th}$ inner-loop update of $\mathbf{F}_{\Phi}$,
\begin{align}
    \theta_{j+1} =
    \begin{cases}
    \theta_{j} - Q_j \nabla_{\theta_{j}}  \mathcal{L}_{\mathcal{D}_{tr}}(f_{\theta_{j}}) \ \ \ \text{if ($j$ mod $w$) $\neq$ $0$,} \\
    (1-P^w_{j})\{\theta_{j} - Q_j \nabla_{\theta_{j}}  \mathcal{L}_{\mathcal{D}_{tr}}(f_{\theta_{j}})\} + P^w_{j} \theta_{j-w} & \text{else}.
    \end{cases}
    \label{eq:update_rule}
\end{align}
If the condition $(j \text{ mod } w )\neq 0$ is satisfied, then the update is simple gradient descent with a preconditioning. If not, it first uses a gradient descent update for the model parameters and then combines the updated weights with previous weight from the $(j-w)^{th}$ step. The current weights and the historical weights are combined with coupling coefficients, $1-P^w_j$ and $P^w_j$, respectively. Similar to $\bm{Q}$, having a per-parameter model for $\bm{P}$ increases the memory complexity by the number of gradient-skip connections. Thus, we use a single meta-parameter set $\bm{P}$ for each layer. This helps to minimize the change in the parameter distribution at a layer.

Note that \ours\ models the complete inner-loop update process (learning trends) by considering how multiple tasks are being optimized. In comparison, Adam~\cite{kingma2014adam} and other adaptive optimizers adapt the learning rate on a single task and for a single step. Ours is also a more general inner-loop function for meta-learning compared to MAML~\cite{finn2017model} and MetaSGD~\cite{li2017meta}. We can recover MAML by setting $w=\infty, Q_j = \mathbf{I}$ and $P = \mathbf{0}$, where $\mathbf{I}$ and $\mathbf{0}$ denote the identity and null {vectors}, respectively. Similarly, we can also recover MetaSGD~\cite{li2017meta} by setting $n=1, w=\infty, Q_j = Q$ and $P^w_j = \mathbf{0}$. See Fig.~\ref{fig:comp} for a visual comparison between the update styles of MAML, Meta-SGD and \ours.

\subsubsection{Outer-loop Optimization} Meta-learning happens in the outer-loop during the optimization process. In the inner-loop, $\theta$ is optimized for a task $\mathcal{T}$, and the objective of the outer-loop is to find a new initialization for $\theta$ and $\Phi$ such that the new tasks can be quickly adapted. Therefore, in the outer-loop, multiple $\theta^k_n$ are available, which are optimized for their corresponding tasks $\mathcal{T}_k\sim {P}(\mathcal{T})$. To combine the knowledge across all the learned tasks, the outer-loop objective is set to minimize the average loss on $\mathcal{D}_{val}^{k}$.
Therefore, we need the gradients with respect to $\theta$ and $\Phi$ on $\mathcal{L}_{\mathcal{D}_{val}}(f_{\theta_n})$,
\begin{align}
    \{ \theta^{new}, \Phi^{new} \} = \{ \theta, \Phi \} - \beta \nabla_{\{\theta, \Phi\}} \sum_{k=1}^K \mathcal{L}_{\mathcal{D}_{val}^k}(f_{\theta^k_n}).
\end{align}
Here, $\beta$ is the learning rate of the outer-loop. Overall, the complete meta-training process can be combined as learning $\theta_n$ using the inner-loop function $\mathbf{F}_{\Phi}$, and learning $\theta$ and $\Phi$ in the outer-loop. Algorithm~\ref{alg:RMAML} outlines the main steps of \ours. The inference stage is represented by Lines 9-13 of Algorithm~\ref{alg:RMAML}. Note that, the outer-loop optimization is similar to MAML; however, the gradients of the meta update are different from MAML and Meta-SGD. This is because of the novel \emph{gradient-skip} connections introduced in the inner-loop function, which bypass the gradients in the backward pass  and combine multiple inner-loop gradients in the forward pass\footnote{Here, backward pass of gradients means the flow of the second-order gradients during the meta update and the forward pass of gradients means first-order gradients for each inner-loop update.}.
It is important to note that `\emph{gradient-skip}' connection denotes bypassing the weights from early updates to the current update, in contrast to the features as in standard residual networks e.g., ResNets.

\begin{algorithm}[t!]
\caption{Meta-training of \ours}
\label{alg:RMAML}
\begin{algorithmic}[1]
\Procedure{Meta training}{}\\
\algorithmicrequire{ $\theta^{init}, w, n, K, \Phi=\{Q_0,.. , Q_{n-1}, P^w_0,.. , P^w_{n-w}\}$}
\State $\theta^0 \gets \theta^{init}$
\For {$i \in [0,1,2,... , n]$ iterations}
    \For {$k \in [1,2,3,... , K]$}
        \State $\theta^i_0 \gets \theta^i$
        \State $\mathcal{T}_k \sim p(\mathcal{T})$
        \State $\mathcal{D}^k_{tr}, \mathcal{D}^k_{val} \gets \mathcal{T}_k$
        \For {$j \in [0,1,2,3,... , n-1]$}
            \State $\theta^i_{j+1} \gets \theta^i_j - Q_j \nabla_{\theta^i_{j}} \mathcal{L}_{\mathcal{D}_{tr}}( f_{\theta^i_j})$
            \If {j mod w == 0 and  j $\geq$ w}\\
                {\qquad\qquad\qquad \footnotesize $\theta^i_{j+1} \gets (1{-}P^w_{j})\{\theta^i_{j} {-} Q_j \nabla_{\theta^i_{j}}  \mathcal{L}_{\mathcal{D}_{tr}}( f_{\theta^i_j})\} {+} P^w_{j} \theta^i_{j-w}$}
            \EndIf
        \EndFor
        \State $\theta^{i,k} \gets \theta^i_n$
    \EndFor
    \State $\theta^{i+1} \gets \theta^i - \beta \nabla_{\theta^i} \sum_{k=1}^K \mathcal{L}_{\mathcal{D}_{val}^k}(f_{\theta^{i,k}})$
    \State $\Phi \ \ \ \gets \Phi - \beta \nabla_{\Phi} \sum_{k=1}^K \mathcal{L}_{\mathcal{D}_{val}^k}(f_{\theta^{i,k}})$
\EndFor
\State \Return $\theta, \Phi$
\EndProcedure
\end{algorithmic}
\end{algorithm}

\begin{algorithm}[t!]
\caption{Online Meta-training}
\label{alg:RMAML}
\begin{algorithmic}[1]
\Procedure{Meta training}{}\\
\algorithmicrequire{ $\theta^{init}, \mathbf{\beta}$}
\State $\theta^1 \gets \theta^{init}$
\State $\phi^1_1 \gets \theta^{1}$
\For {$t \in [1,2,... , T]$ iterations}
    \State get dataset $D_t$
    \For {$i \in [1,2,3,... , N]$}
        \If {i mod k  $\neq$ 0}
            \State $\phi^t_i \gets \phi^t_i - \alpha_1 \nabla_{\phi} \{ \mathcal{L}_t (\phi) + \mathcal{R}(\phi, \theta)\}$
        \Else
            \State add $D_t^i$ to the buffer $\mathbf{\beta}$
            \State $\theta^t \gets \theta^t - \alpha_2 \nabla_{\theta} \{ \phi - \nabla_{\theta} \{ \mathcal{L}_t (\phi) + \mathcal{R}(\phi, \theta)\}$
        \EndIf
    \EndFor
\EndFor
\EndProcedure
\end{algorithmic}
\end{algorithm}

\subsection{Gradients Analysis}

In \ours, there exist two types of preconditioning. The first preconditioning happens at each update step, with $Q_j$ transforming the forward gradient into the preconditioned direction. Here, the meta-parameters $Q_j$ are kernel-wise normalized vectors. Therefore, the main role of $Q_j$ is to learn a better direction to project the true gradients towards the globally optimal parameter manifold.

\begin{figure}[!t]
    \begin{minipage}[t]{.5\textwidth}
    \centering
    \includegraphics[angle=-90, scale=0.30]{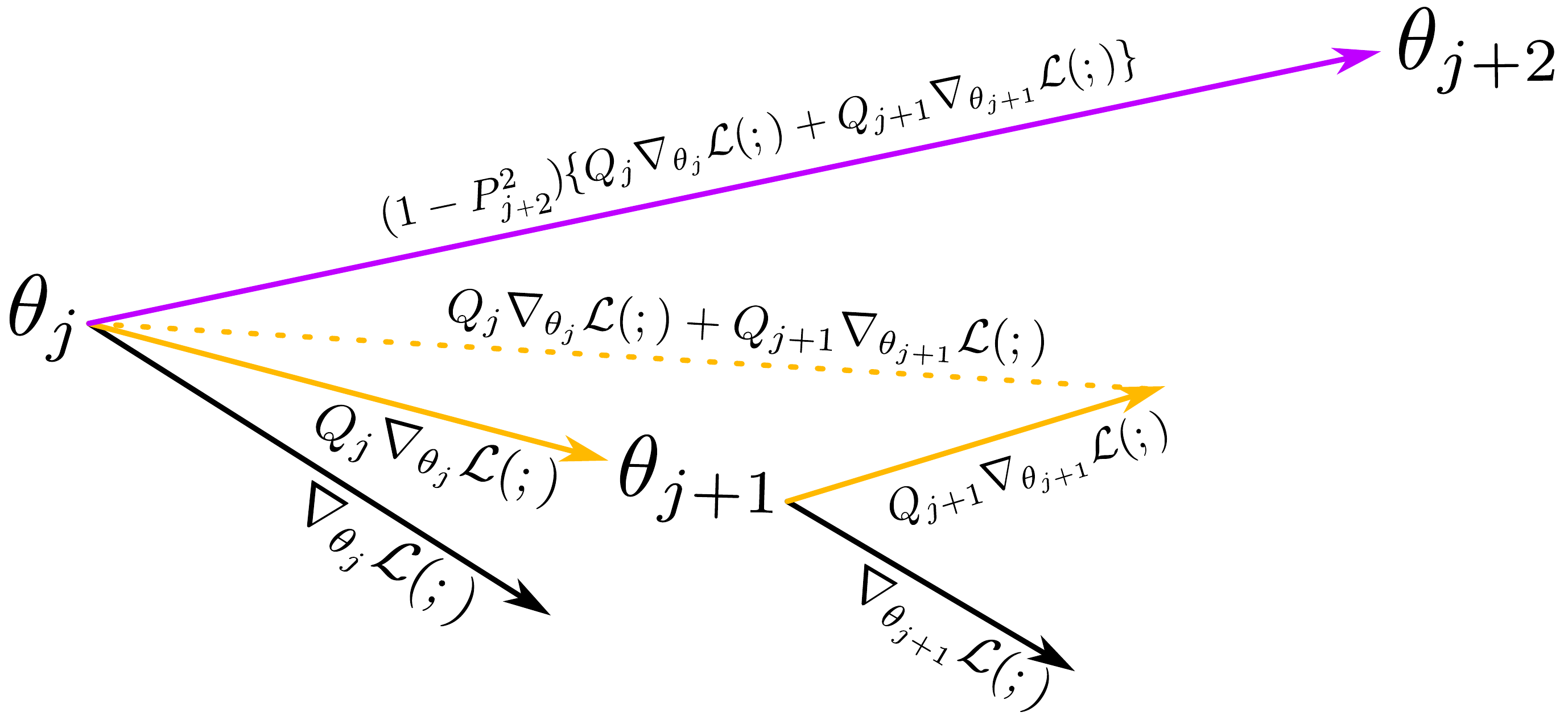}
    \end{minipage} \hfill
    \begin{minipage}[t]{.5\textwidth}
    \caption{\emph{Gradient flow in the inner-loop updates of \ours \ with $w=2$:} The true gradients are shown in black, and $Q_j, Q_{j+1}$ are used to precondition these gradients (in yellow). $P_{j+2}^2$ is used to combine the preconditioned gradients of the update steps $j, j+1$.  }
    \label{fig:grad}
    \end{minipage}
\end{figure}

In addition to learning the unique trends in each step, \ours\ has multiple gradient-skip connections in the inner-loop function $\mathbf{F}_{\Phi}$. These connections allow our model to learn from the previous updates. Consider a gradient-skip connection with a coefficient $P^w_j$, where $w$ is the interval length, and $j-w$ is the starting point of the gradient-skip connection. Therefore,
the role of the $P^w_j$ is to learn  a relationship between the updates at the $j^{th}$ and $(j-w)^{th}$  steps and project a better preconditioned gradient for the $j^{th}$ update, implicitly encapsulating all the gradient updates between $j$ and $j-w$ steps.
\begin{align*}
    \theta_{j+1} &= (1-P^w_j) \{ \theta_{j} - Q_{j}\nabla_{\theta_{j}} \mathcal{L}_{\mathcal{D}_{tr}}(f_{\theta_{j}}) \} + P^w_j \theta_{j-w} \\
    &= (1-P^w_j) \{ \theta_{j-1} - Q_{j-1}\nabla_{\theta^i_{j-1}} \mathcal{L}_{\mathcal{D}_{tr}}(f_{\theta_{j-1}})  \\
    & \ \ \ \ \ \ \ \ \  \ \ \ \ \ \ \ \ \ \ \ \ \ \ \ - Q_{j}\nabla_{\theta_{j}} \mathcal{L}_{\mathcal{D}_{tr}}(f_{\theta_{j}}) \} + P^w_j \theta_{j-w} \\
    &= \theta_{j-w} - (1-P^w_j) \bigg\{\sum_{s=0}^{w} Q_{j-s}\nabla_{\theta_{j-s}} \mathcal{L}_{\mathcal{D}_{tr}}(f_{\theta_{j-s}}) \bigg\}.
\end{align*}
We note that, for the $(j+1)^{th}$ update, all the previous gradients upto last $w$ updates are combined and preconditioned to generate the final gradient. See Fig.~\ref{fig:grad} for the flow of the meta-transformed gradients for $w=2$.

During the meta updates, we back-propagate through the inner-loop learning process $\mathbf{F}_{\Phi}$. The meta gradients are computed over the average validation loss on each task. These meta gradients encapsulate information across tasks, making it possible to `learn about learning'. If we decompose the meta-gradient $\mathcal{G}_{\ours}$, we can see that the gradients are bypassed to the early steps due to the existence of gradient-skip connections in our update process. For the sake of brevity, we consider the setting where $n \text{ mod } w=0$ and $Q_j=\mathbf{I}$. However, the following analysis can easily be extended to the rest of the cases without loss of generality,
\begin{align}
    & \mathcal{G}_{\ours} = \nabla_{\theta} \mathcal{L}_{\mathcal{D}_{val}}(f_{\theta_n})
    = \underbrace{\nabla_{\theta_n} \mathcal{L}_{\mathcal{D}_{val}}(f_{\theta_n})}_{G} \cdot \nabla_{\theta} \theta_n, \\
    &= G \cdot \nabla_{\theta} \bigg\{ \theta - \underbrace{\sum_{r=0}^{n\,\floordiv\,w} (1-P^w_r)}_{\widehat{P}} \underbrace{\sum_{s=0}^{w-1} \nabla_{\theta_{s+rw}}\mathcal{L}_{\mathcal{D}_{tr}}(f_{\theta_{s+rw}})}_{\widehat{\nabla}} \bigg\}, \notag \\
    &= G \cdot \nabla_{\theta} \bigg\{ \theta \big(1 - \widehat{P}\big) + \widehat{P} \left\{ \theta - \widehat{\nabla} \right\} \bigg\}, \notag\\
    &=  G \cdot (1-\widehat{P}) + \underbrace{G \big( 1 - \sum_{r=0}^{n \,\floordiv\,w} \nabla_{\theta}\widehat{\nabla} \big)}_{\text{MAML gradient}} - \underbrace{G \sum_{r=0}^{n\,\floordiv\,w} P^w_r \big( 1 -  \nabla_{\theta}\widehat{\nabla} \big)}_{\mu}, \notag \\
    &= G \cdot (1-\widehat{P}) + \mathcal{G}_{MAML} - \mu  \approx G \cdot (1-\widehat{P}) + \mathcal{G}_{MAML}. \notag
\end{align}
The above analysis shows that during the  back-propagation of meta-update in MAML, the gradients on the validation data at the final iteration $G$ are transformed by multiple second-order gradients at each update. Hence, the gradients could vanish by the time they  arrive at the first update point during back-propagation. However, in \ours, the meta-gradient is composed of the MAML gradient and an additional gradient term without containing any second-order derivatives. Due to the gradient-skip connections, this term is bypassed to the meta-update, which helps mitigate gradient vanishing. Thus, \ours can be trained for many inner-loop iterations without overfitting on a given task.

\section{Experiments}

We extensively evaluate \ours on few-shot supervised classification and regression problems. In both cases, the training set has a large number of tasks with a very small number of samples. The fewer samples per task makes it non-trivial to learn a good representation using traditional supervised learning approaches. However, since there are many tasks available, we can learn `what is the best way to learn a task'. Therefore, meta-learning frameworks are well suited for few-shot learning problems. In addition to evaluating on few-shot learning, we provide a detailed ablation study on \ours (Sec.~\ref{sec: ablation}).

\subsection{Few-shot Classification}

\noindent\textbf{Settings:} For an $M$-way $N$-shot setting, $N$ samples are given for $M$ different classes in each task. Therefore, for a $k^{th}$ task $\mathcal{T}_{k}$, the training dataset $\mathcal{D}_{tr}^k \in \mathcal{T}_{k}$ has $N*M$ samples. For all experiments, we use $15$ samples/class for validation during the meta-update, hence $\mathcal{D}_{val}^k \in \mathcal{T}_{k} $ has $M * 15$ samples.

\noindent\textbf{Datasets:} We evaluate on three benchmarks for few-shot classification. \textbf{(a)} miniImageNet~\cite{ravi2016optimization} has 100 classes, and is a subset of ImageNet. We use the same split as in~\cite{ravi2016optimization}, with 64, 16 and 20 classes for train, validation and test sets.  \textbf{(b)} CIFAR-FS~\cite{bertinetto2018meta} has 100 object classes, split into 64, 16, and 20 for training, validation and test sets, respectively.
\textbf{(c)} tieredImageNet~\cite{ren2018meta} is also a subset of ImageNet, with 608 classes grouped into 34 high-level categories, divided into 20, 6 and 8 for training, validation, and testing.

\noindent\textbf{Implementation Details:} For experiments on miniImageNet, CIFAR-FS and tieredImageNet, we use a simple four-layer convolutional neural network, each with 64  filters. During the meta update we use 15 samples per class in the validation data  to compute the meta gradients (lines  14-15 in Algorithm~\ref{alg:RMAML}) and batch size of four tasks ($k$=4 in Algorithm~\ref{alg:RMAML}) used in all the experiments. For all the experiments, the inner-loop is updated five times ($n$=5 in Algorithm~\ref{alg:RMAML}) with two gradient-skip connections using $\bm{P}^2$. Also, the inner-loop learning rate is set to 0.01 and, in the outer-loop, Adam~\cite{kingma2014adam} optimizer with an initial learning rate of 0.001 is employed. All the models are updated for up to $60{,}000$ iterations ($i$=60k in Algorithm~\ref{alg:RMAML}) and the hyper-parameters are set as in~\cite{finn2017model}, for a fair comparison. Also, we used a random crop and random flip augmentations for miniImageNet and CIFAR-FS.
Cross-entropy loss is used as the task objective as well as meta-objective, on the $\mathcal{D}_{tr}$ and $\mathcal{D}_{val}$, respectively.

\begin{table}[!t]
    \centering
            \begin{center}
\setlength{\tabcolsep}{0.3cm}
\begin{tabular}{l c c}
\toprule[0.4mm]
\multirow{2}{*}{Methods} & \multicolumn{2}{c}{miniImageNet, 5-way}  \\
& 1-shot & 5-shot \\ \midrule
MAML~\cite{finn2017model}            & 48.70 $\pm$ 1.84 & 63.11 $\pm$ 0.92  \\
MAML++~\cite{antoniou2018train}      & 52.15 $\pm$ 0.26  & 68.32 $\pm$ 0.44  \\
FOMAML~\cite{finn2017model}          & 48.07 $\pm$ 1.75  & 63.15 $\pm$ 0.91  \\
Reptile~\cite{nichol2018first}       & 49.97 $\pm$ 0.32  & 65.99 $\pm$ 0.58  \\
Meta-LSTM~\cite{ravi2016optimization} & 43.44 $\pm$ 0.77  & 60.60 $\pm$ 0.71  \\
Meta-SGD~\cite{li2017meta}           & 50.47 $\pm$ 1.87  & 64.03 $\pm$ 0.94  \\
iMAML-HF~\cite{rajeswaran2019meta}   & 49.30 $\pm$ 1.88  & - \\
MT-Net~\cite{mtnet}                  & 51.70 $\pm$ 1.84  & - \\
R2D2~\cite{bertinetto2018meta}       & 49.50  $\pm$ 0.20   & 65.40 $\pm$ 0.20  \\
L-MAML~\cite{finn2018metalearning}   & 49.40 $\pm$ 1.83  & - \\
HSML~\cite{Hierarchically}           & 50.38 $\pm$ 1.85  & - \\ \midrule
\ours                                &   \textbf{53.50 $\pm$ 0.89 } & \textbf{70.51 $\pm$ 0.67 } \\
\bottomrule[0.4mm]
\end{tabular}
\end{center}
\caption{Few-shot learning results on miniImageNet~\cite{ravi2016optimization} dataset. All accuracies are in \%.}

\label{tbl:results1}
\end{table}

\begin{table}[h]
    \begin{center}
    \resizebox{1\columnwidth}{!}{
        \setlength{\tabcolsep}{0.08cm}
        \begin{tabular}{l c c |c c}
\toprule[0.4mm]
\multirow{2}{*}{Methods} & \multicolumn{2}{c|}{CIFAR-FS, 5-way} & \multicolumn{2}{c}{tieredImageNet, 5-way} \\
& 1-shot & 5-shot &  1-shot & 5-shot  \\ \midrule
Meta-LSTM~\cite{ravi2016optimization}    &  43.4 $\pm$ 0.8    &  60.6 $\pm$ 0.7   &  - &  - \\
MAML~\cite{finn2017model}                        &  58.9 $\pm$ 1.9    &  71.5 $\pm$ 1.9   &  51.67 $\pm$ 1.87   &  70.30 $\pm$ 1.75   \\
Meta-SGD~\cite{li2017meta}                       &  56.9 $\pm$ 0.9    &  70.1 $\pm$ 0.7   &  50.92 $\pm$ 0.93   &  69.28 $\pm$ 0.80   \\
R2D2~\cite{bertinetto2018meta}                   &  62.3 $\pm$ 0.2    &  77.4 $\pm$ 0.2   &  -   &  -  \\ \midrule
\ours                         &    \textbf{63.5 $\pm$ 0.9 }   & \textbf{79.1 $\pm$ 0.7 }  & \textbf{54.81 $\pm$ 0.88 }  &  \textbf{74.39 $\pm$ 0.71}  \\
\bottomrule[0.3mm]
\end{tabular} }
\end{center}
\caption{Few-shot learning results on CIFAR-FS~\cite{bertinetto2018meta} and tieredImageNet~\cite{ren2018meta} datasets. All accuracies are in \%. }
\label{tbl:results1}
\end{table}

\begin{figure*}[!h]
    \centering
\begin{minipage}{0.7\textwidth}
    \begin{subfigure}{0.5\textwidth}
    \label{fig:meta_agnostic}
        \centering
        \includegraphics[width=1\textwidth,clip=true,trim=1.5cm 1.5cm 1.5cm 1.5cm]{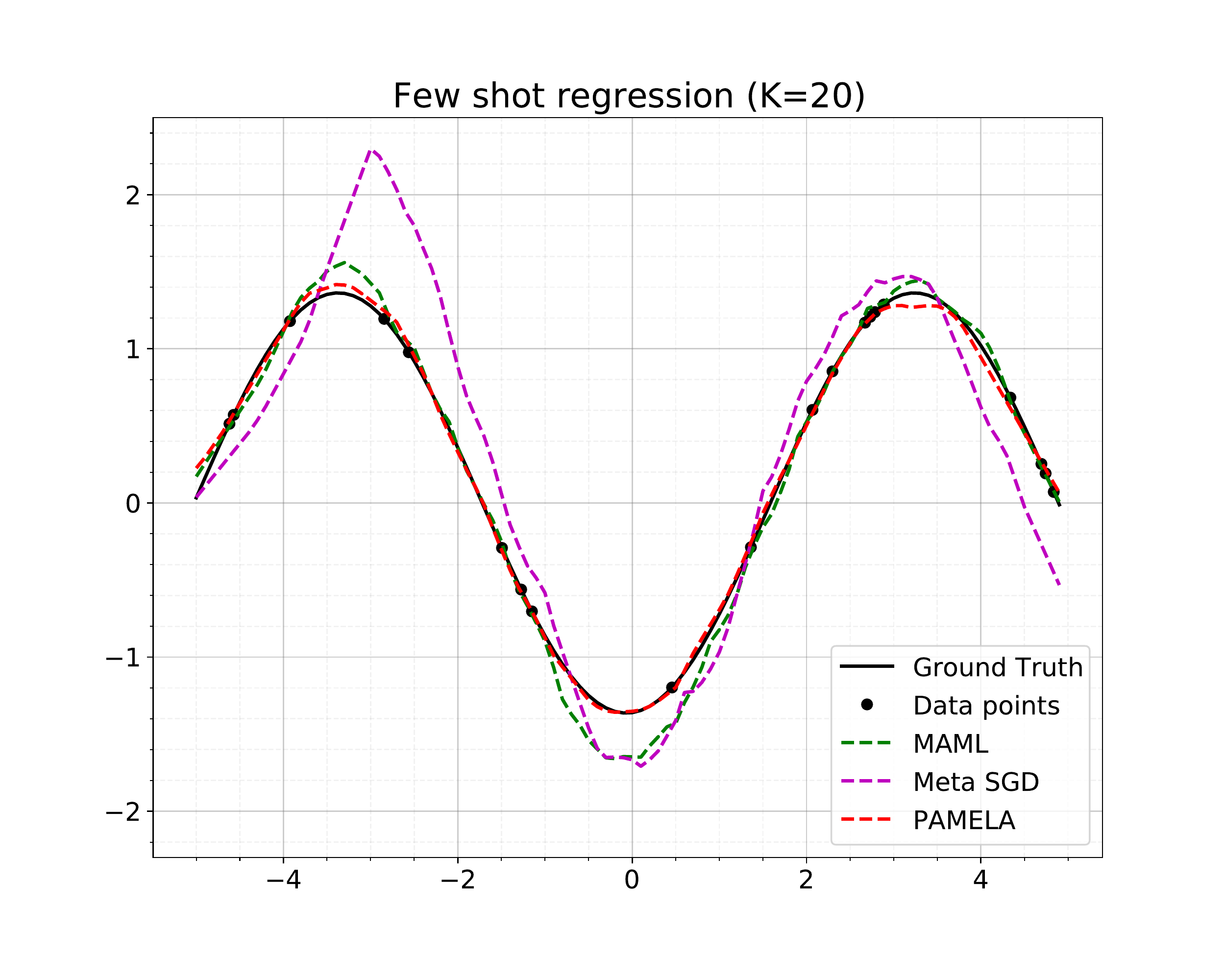}
    \end{subfigure}%
    \begin{subfigure}{0.5\textwidth}
        \centering
        \includegraphics[width=1\textwidth,clip=true,trim=1.5cm 1.5cm 1.5cm 1.5cm]{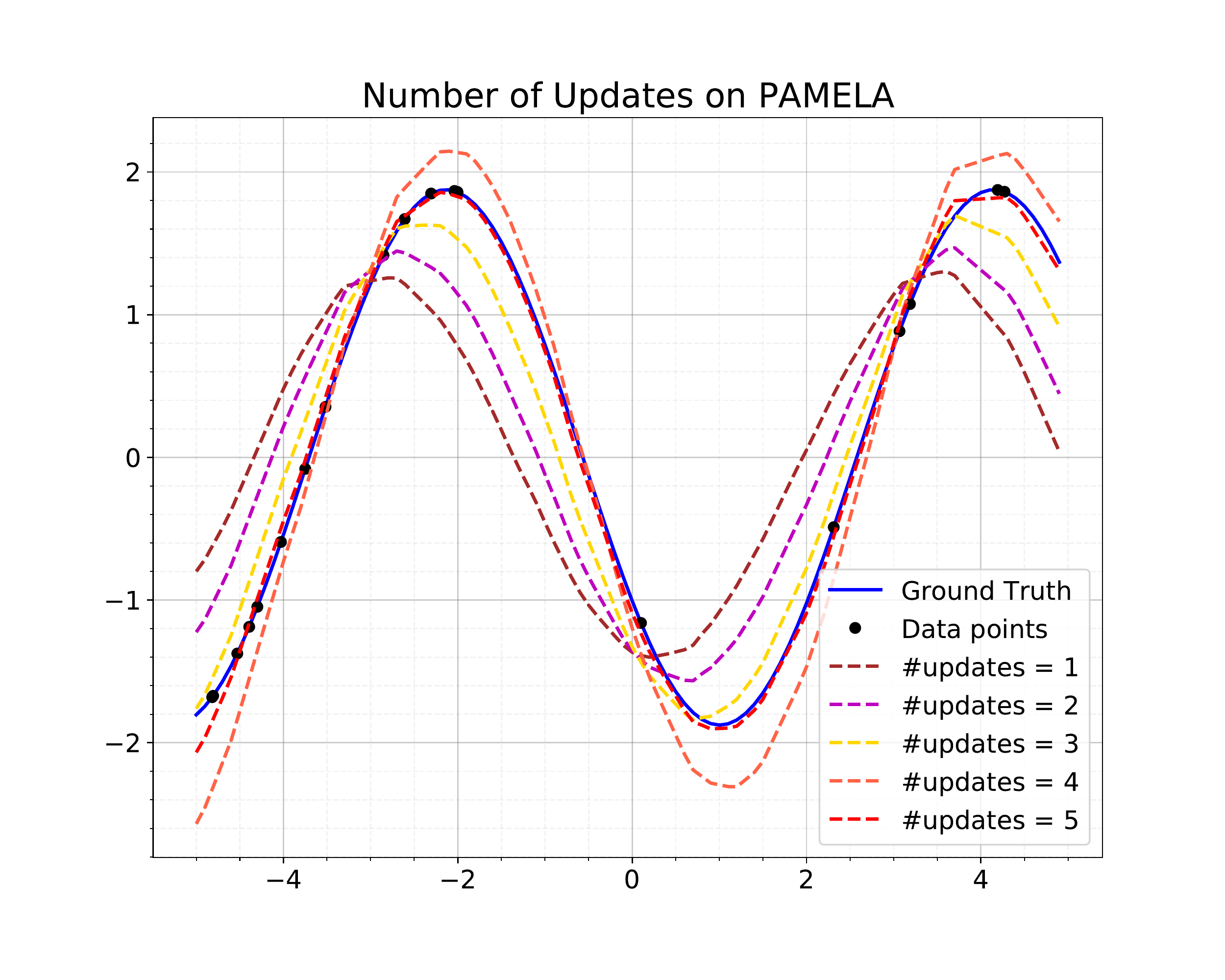}
    \end{subfigure}%
    \setlength{\belowcaptionskip}{-10pt}
    \captionof{figure}{\emph{Sine wave regression:} \textbf{left:} We compare \ours with MAML and Meta-SGD on the sine wave regression problem. We can see that \ours \ fits the curve better, and the curve generated by \ours\ is smoother than the curve generated by MAML and Meta-SGD. \textbf{right:} We also show that how the function changes with the number of inner-loop updates. Note that, at 4th update, the model over shoots, however in the next update it recovers the better function.}
    \label{fig:sine}
\end{minipage}
\hfill
\begin{minipage}{0.29\textwidth}
    \centering\setlength{\tabcolsep}{8pt}
    \scalebox{0.8}{
    \begin{tabular}{c l c}
                \toprule[0.4mm]
              K &  Method & MSE \\ \midrule
            \multirow{3}{*}{5}  &   MAML~\cite{finn2017model}           & 1.13 $\pm$ 0.18 \\
                                &   Meta-SGD~\cite{li2017meta}          & 0.90 $\pm$ 0.16 \\
                                &   MT-Net~\cite{mtnet}                 & 0.76 $\pm$ 0.09 \\
                                &   \ours                               & 0.54 $\pm$ 0.06 \\  \midrule
            \multirow{3}{*}{10} &   MAML~\cite{finn2017model}           & 0.77 $\pm$ 0.11 \\
                                &   Meta-SGD~\cite{li2017meta}          & 0.53 $\pm$ 0.09 \\
                                &   MT-Net~\cite{mtnet}                 & 0.49 $\pm$ 0.05 \\
                                &   \ours                               & 0.41 $\pm$ 0.04 \\  \midrule
            \multirow{3}{*}{20} &   MAML~\cite{finn2017model}           & 0.48 $\pm$ 0.08 \\
                                &   Meta-SGD~\cite{li2017meta}          & 0.31 $\pm$ 0.05 \\
                                &   MT-Net~\cite{mtnet}                 & 0.33 $\pm$ 0.04 \\
                                &   \ours                               & 0.17 $\pm$ 0.03 \\
                \bottomrule[0.4mm]
    \end{tabular}
    }
      \captionof{table}{\emph{Mean squared error on sine wave regression:} We test MAML-based algorithms on sine wave regression with $K=\{5,10,20\}$. \ours\ achieves the lowest error in all three cases. \emph{(lower is better)} }
      \label{tbl:mse}
\end{minipage}
\end{figure*}

\begin{figure}[!h]
    \centering
    \begin{minipage}{0.39\columnwidth}
        \captionof{figure}{Loss on different steps of the inner-loop updates. We can see that \ours\ converges with decreasing variation in error compared to MAML and does not overfit with the increasing number of inner-loop updates. Losses were calculated across 100 runs.}
        \label{fig:loss}
    \end{minipage}
    \hfill
    \begin{minipage}{0.6\columnwidth}
       \includegraphics[scale=0.32]{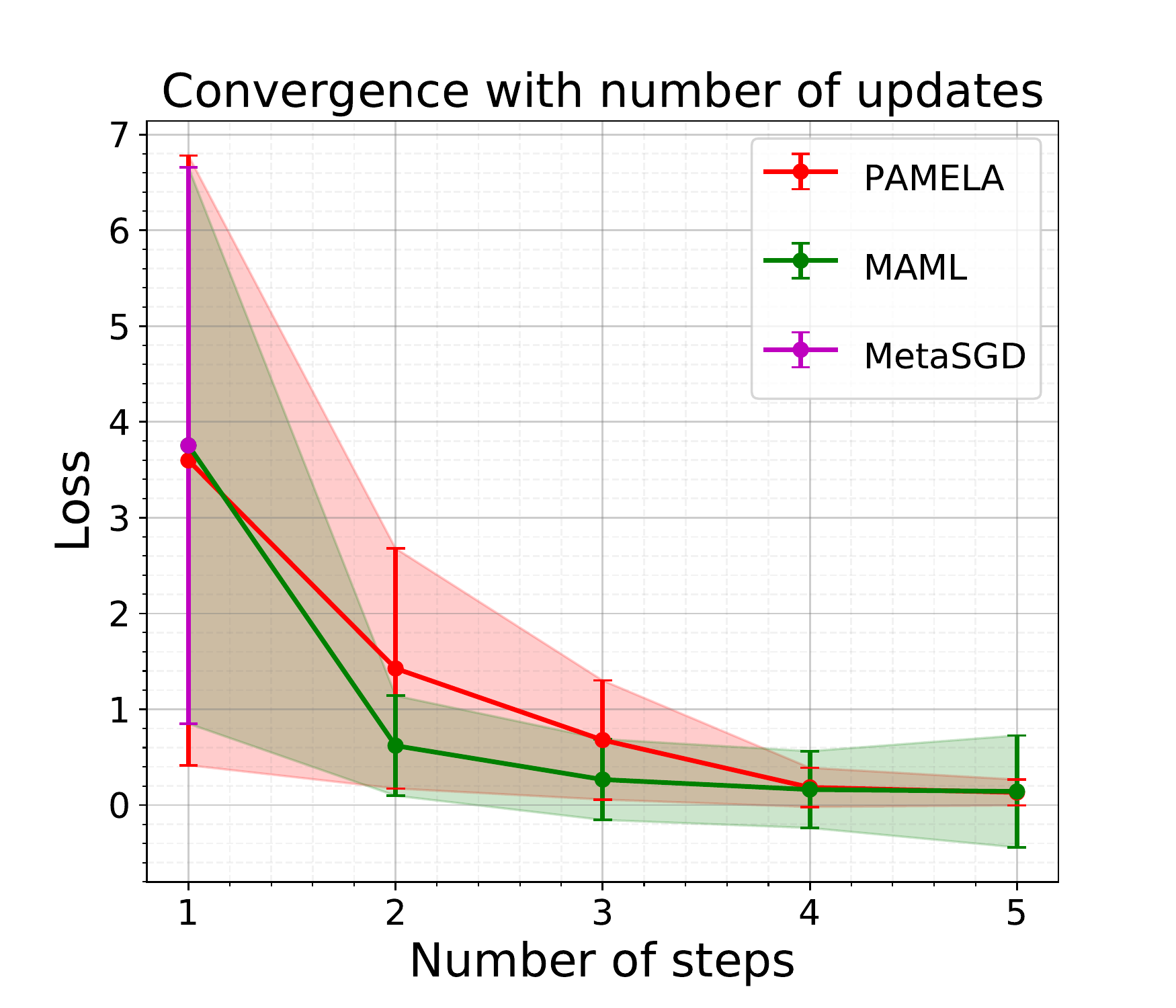}
    \end{minipage}
\end{figure}

\noindent\textbf{Results:} We compare \ours\ with several other model-agnostic gradient-based methods: MAML~\cite{finn2017model}, iMAML~\cite{rajeswaran2019meta}, FOMAML~\cite{finn2017model}, Reptile~\cite{nichol2018first} and Meta-SGD~\cite{li2017meta}. On the miniImagenet 5-way 5-shot setting, with a total of $600$ test tasks (out of possible $15{,}504$ tasks), \ours achieves $70.51 \pm 0.67\%$, whereas for the 5-way 1-shot setting, it achieves $53.50 \pm 0.89\%$. This is an absolute gain of $2.19\%$ and $1.35\%$ over the current state-of-the-art method. Similarly, on the CIFAR-FS dataset, our method achieves state-of-the-art performance compared to the family of MAML algorithms. On the 5-way 5-shot setting, we achieve $79.1 \pm 0.7\%$, with a gain of $1.7\%$, and on 5-way 1-shot setting, we achieve $63.5 \pm 0.9\%$ with a gain of $1.2\%$. Further, we also surpass MAML, Meta-SGD on  5- and 1-shot learning on tieredImageNet with $54.81 \pm 0.88\%$ and $74.39 \pm 0.71\%$. The path learning behaviour of PAMELA provides valuable context and gradient preconditioning, resulting in strong improvements compared to other model-agnostic gradient-based methods.

\subsection{Regression}

\textbf{Settings:} We consider sine wave regression as our learning problem. Each task/wave is parameterized by amplitude, frequency and phase sampled uniformly in ranges $[0.1,5]$, $[0.8,1.2]$ and $[0,\pi]$, respectively. In addition, a task $\mathcal{T}$ contains $K + 10$ samples uniformly drawn from the true curve. The first $K$ samples are used for the inner-loop adaptation and the remaining $10$ used to compute the meta-loss.

\noindent\textbf{Implementation Details:} For the regression setting, we use a simple multi-layer perception with two hidden layers, each with $40$ ReLU neurons. We follow the same update rule as in Eq.~\ref{eq:update_rule} and cross-entropy is replaced with $L_2$ loss.

\noindent\textbf{Results:} To evaluate the effectiveness of the model, we first sample 1000 random sine waves from the same distribution as the training set. Then we sample $K$ random points on the curve to update the model. Afterwards, we sample $1000$ points on the curve uniformly to find the Mean Squared Error (MSE) to evaluate the fitness of the curve. We test on MAML, Meta-SGD, MT-Net and \ours\ with $K=\{5,10,20\}$. In all the cases, our method obtains the lowest error, and with increasing $K$ the loss decreases faster (see Table~\ref{tbl:mse}). From Fig.~\ref{fig:sine}, we can see that the curves regressed by \ours\ fit the target function well, and are much smoother compared to other methods. Additionally, Fig.~\ref{fig:loss} shows how the loss changes with the number of inner-loop updates. We can see that, \ours\ has better convergence compared to MAML (standard deviation of loss decreases with inner-loop updates).

\subsection{Ablation Studies}
\label{sec: ablation}

\textbf{Dissecting \ours:} We analyze individual contributions of different components of \ours. We can breakdown \ours\ as a mixture of MAML + $\bm{Q}$ + $\bm{P}^w$. First, we see how $\bm{Q}$ affects the model learning curve. If we use only one $Q$ for all the inner-loop updates, extending Meta-SGD, the performance degrades. This is because $Q$ acts as a learning rate controller for each step, while having the same learning rates for all the steps, may cause the model to over-shoot or prevent it from converging. However, using different $Q_j$ for each update helps to boost the performance by $3.07\%$. In addition to $\bm{Q}$, $\bm{P}^w$ helps to learn the trend across different updates. Therefore, adding $\bm{P}^w$ to the loop makes the learning more generic. As a result, the gradient-skip connection ($\bm{P}$) gives the highest boost in the performance, with a $7.40\%$ gain over MAML. However, $\bm{P}^w$ contains $P^w_j$ with a hyper-parameter of interval size ($w$), which controls the gap between each gradient-skip connection. We can see how the length of the gradient-skip connections affects the learning from Table~\ref{tbl:ab1}. Our results show that the length of the skip connection is not very  significant.

\begin{table}[!h]
\footnotesize
    \centering
            \begin{center}
            \begin{tabular}{l c}
\toprule[0.4mm]
Method & Accuracy (\%) \\
\midrule
MAML                                 & 63.11 $\pm$ 0.92 \\
MAML + $Q_0$ (single step)                & 64.03 $\pm$ 0.94  \\
MAML + $Q_0$ (multiple steps)           & 62.71 $\pm$ 0.69  \\
MAML + $\bm{Q}$                      & 69.71 $\pm$ 0.66  \\
MAML + $\bm{P}$                      & 68.30 $\pm$ 0.67  \\ \midrule
MAML + $\bm{Q}$ + $\bm{P}^1$         & 70.72 $\pm$ 0.63  \\
MAML + $\bm{Q}$ + $\bm{P}^2$         & 70.17 $\pm$ 0.67  \\
MAML + $\bm{Q}$ + $\bm{P}^3$         & 70.31 $\pm$ 0.69  \\
MAML + $\bm{Q}$ + $\bm{P}^4$         & 70.51 $\pm$ 0.65  \\
\bottomrule[0.4mm]
\end{tabular}
\end{center}
\captionof{table}{\emph{Dissecting \ours}: Results are reported for miniImageNet 5-way 5-shot experiments with 600 testing samples, and $n=5$. Means and 95\% confidence intervals are reported. Although, the individual use of multiple $Q_j$ and gradient skip connections help in modelling a multi-step adaptation process, combination of both gives an overall better performance.  }
\label{tbl:ab1}
\end{table}

\noindent\textbf{Computational Complexity of \ours:} Due to the extensive modeling of  inner-loop paths, the memory and computational complexity of \ours\ is higher than MAML. However, \ours\ is designed to make sure this overhead is minimum. For example, the training time and test test time for a task is only about 5\% higher than of MAML (see Table~\ref{tbl:complexity}). Note that, both MAML and ours use 5 inner-loop updates, while Meta-SGD uses a single update. Additionally, ours has only about 33\% more parameters than MAML, while Meta-SGD has 100\% additional parameters.

\begin{table}[h]
\footnotesize
    \centering
            \begin{center}
            \begin{tabular}{l c c r r}
\toprule[0.4mm]
Method & \#Parameters & \#Updates &  Train time & Test time \\
\midrule
MAML                    & 121.09$\times 10^3$ &  5 & 205.50 ms  &   41.33 ms \\
MetaSGD                 & 242.18$\times 10^3$ &  1 &  53.75 ms  &  16.67 ms \\
PAMELA                  & 162.39$\times 10^3$ &  5 &  215.84 ms  &  43.33 ms \\
\bottomrule[0.4mm]
\end{tabular}
\end{center}
\caption{\emph{Model complexity:} We compare the number of parameters, train time per task and test time per task in a 4-layer 64-channel CNN. Note that MetaSGD allows a single inner-loop update.}
\label{tbl:complexity}
\end{table}

\begin{figure*}[t!]
    \centering
        \centering
        \includegraphics[width=0.98\textwidth]{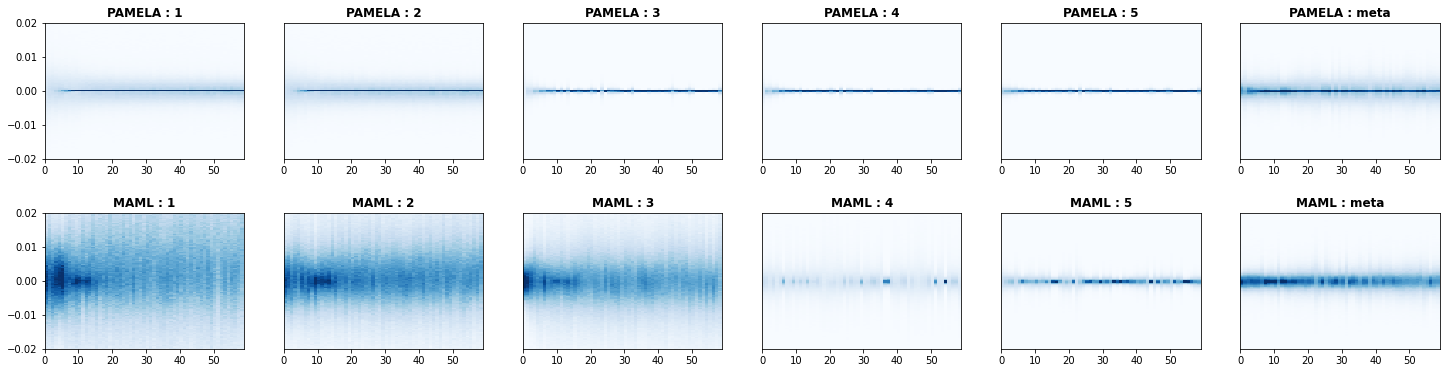}
    \setlength{\belowcaptionskip}{-8pt}
    \caption{\emph{Inner-loop and outer-loop gradients for the final layer (conv4):} (left to right) First 5 columns represent the inner-loop gradients from 1-5 steps, and the last column represents the meta-gradients. In each plot, the horizontal axis is for the number of epochs and the vertical axis indicates the gradient magnitude. After each training episode, the gradients are grouped into 512 bins and number of items in a bin is proportional to the strength of blue color in the plot.}
    \label{fig:grad2}
\end{figure*}

\noindent\textbf{Number of inner-loop updates and skip connections:} Our inner-loop is modeled with two hyper-parameters: the number of updates and the width of the skip connections. First, as we change the number of inner-loop updates, we notice the performance does not change significantly except for the single update scenario. This is mainly because the model's inner-loop path has an extra degree of freedom on how far the parameters have to move. While training the model, it can learn how the inner-loop path has to change for a given constraint on the total updates. All these experiments were done on miniImageNet with 5-way 5-shot problems with $\bm{P}^2$ gradient-skip connections. Secondly, the gradient-skip connection can also change the dynamics of the adaptation. Although it helps to have the gradient skip connections, the width of the connections does not have a significant impact on the performance. However, a depth of 5-10 updates may not be sufficient enough to prove this claim.

\noindent\textbf{Diversity of the gradients:} The inner-loop dynamics of \ours\ are different from MAML due to the gradient preconditioning and gradient-skip connections. To validate this, we visualize the histogram of gradients in 
\textit{conv4} layer after each training iteration (see Fig.~\ref{fig:grad2}).
The plots are shown for first 5 inner-loop updates and meta-update for both \ours\ and MAML.
One can note that the magnitude of gradients for \ours\ decreases as training progress. In comparison, the meta-gradients of MAML are about one order of magnitude higher than \ours. Among the inner-loop gradients, \ours\ converges faster towards minimal (close to zero) updates, while MAML requires large magnitude of gradients till last update. For the meta updates, while ours stays around zero, MAML gradients diverge and increase in magnitude as we progress. This indicates that the convergence achieved by MAML towards a better initialization is perhaps sub-optimal as it requires large gradients to adjust the parameters. Small inner-loop gradients means a little adjustment in parameters is sufficient to get to the task-specific model, hence a better initialization.

\section{Conclusion}
Model-agnostic meta-learning aims to combine across-task knowledge in order to find a better initialization for the learner. From this initialization, the model can be quickly fine-tuned on a new task with only a few examples. In the existing models, the relationships among the tasks with respect to a task-specific (inner-loop) learning path are not modeled. In this work, we propose to learn a gradient preconditioning at each inner-loop iteration to model how learning evolves over time across multiple tasks. Further, our approach utilizes historical gradients from old updates, which provides valuable context and prevents over-fitting as well as gradient vanishing. Overall, our approach achieves faster convergence and performs better classification datasets, including miniImageNet, tieredImageNet, and CIFAR-FS. We also evaluate our approach on a sine wave regression task, where it fits the curve much more accurately than the state-of-the-art methods. Further, we analyse the contribution of each respective component towards the final performance via an extensive ablation study.

\ifCLASSOPTIONcaptionsoff
  \newpage
\fi

\bibliographystyle{IEEEtran}
\bibliography{egbib}

\begin{thebibliography}{10}
\providecommand{\url}[1]{#1}
\csname url@samestyle\endcsname
\providecommand{\newblock}{\relax}
\providecommand{\bibinfo}[2]{#2}
\providecommand{\BIBentrySTDinterwordspacing}{\spaceskip=0pt\relax}
\providecommand{\BIBentryALTinterwordstretchfactor}{4}
\providecommand{\BIBentryALTinterwordspacing}{\spaceskip=\fontdimen2\font plus
\BIBentryALTinterwordstretchfactor\fontdimen3\font minus
  \fontdimen4\font\relax}
\providecommand{\BIBforeignlanguage}[2]{{%
\expandafter\ifx\csname l@#1\endcsname\relax
\typeout{** WARNING: IEEEtran.bst: No hyphenation pattern has been}%
\typeout{** loaded for the language `#1'. Using the pattern for}%
\typeout{** the default language instead.}%
\else
\language=\csname l@#1\endcsname
\fi
#2}}
\providecommand{\BIBdecl}{\relax}
\BIBdecl

\bibitem{finn2017model}
C.~Finn, P.~Abbeel, and S.~Levine, ``Model-agnostic meta-learning for fast
  adaptation of deep networks,'' in \emph{International Conference on Machine
  Learning}, 2017, pp. 1126--1135.

\bibitem{li2017meta}
Z.~Li, F.~Zhou, F.~Chen, and H.~Li, ``Meta-sgd: Learning to learn quickly for
  few-shot learning,'' \emph{arXiv preprint arXiv:1707.09835}, 2017.

\bibitem{park2019meta}
E.~Park and J.~B. Oliva, ``Meta-curvature,'' in \emph{Advances in Neural
  Information Processing Systems}, 2019, pp. 3309--3319.

\bibitem{antoniou2018train}
A.~Antoniou, H.~Edwards, and A.~Storkey, ``How to train your maml,''
  \emph{arXiv preprint arXiv:1810.09502}, 2018.

\bibitem{hochreiter2001learning}
S.~Hochreiter, A.~S. Younger, and P.~R. Conwell, ``Learning to learn using
  gradient descent,'' in \emph{International Conference on Artificial Neural
  Networks}.\hskip 1em plus 0.5em minus 0.4em\relax Springer, 2001, pp. 87--94.

\bibitem{andrychowicz2016learning}
M.~Andrychowicz, M.~Denil, S.~Gomez, M.~W. Hoffman, D.~Pfau, T.~Schaul,
  B.~Shillingford, and N.~De~Freitas, ``Learning to learn by gradient descent
  by gradient descent,'' in \emph{Advances in Neural Information Processing
  Systems}, 2016, pp. 3981--3989.

\bibitem{ravi2016optimization}
S.~Ravi and H.~Larochelle, ``Optimization as a model for few-shot learning,''
  \emph{International Conference on Learning Representations}, 2017.

\bibitem{koch2015siamese}
G.~Koch, R.~Zemel, and R.~Salakhutdinov, ``Siamese neural networks for one-shot
  image recognition,'' in \emph{ICML deep learning workshop}, vol.~2.\hskip 1em
  plus 0.5em minus 0.4em\relax Lille, 2015.

\bibitem{vinyals2016matching}
O.~Vinyals, C.~Blundell, T.~Lillicrap, D.~Wierstra \emph{et~al.}, ``Matching
  networks for one shot learning,'' in \emph{Advances in Neural Information
  Processing Systems}, 2016, pp. 3630--3638.

\bibitem{sung2018learning}
F.~Sung, Y.~Yang, L.~Zhang, T.~Xiang, P.~H. Torr, and T.~M. Hospedales,
  ``Learning to compare: Relation network for few-shot learning,'' in
  \emph{IEEE Conference on Computer Vision and Pattern Recognition}, 2018, pp.
  1199--1208.

\bibitem{snell2017prototypical}
J.~Snell, K.~Swersky, and R.~Zemel, ``Prototypical networks for few-shot
  learning,'' in \emph{Advances in Neural Information Processing Systems},
  2017, pp. 4077--4087.

\bibitem{santoro2016meta}
A.~Santoro, S.~Bartunov, M.~Botvinick, D.~Wierstra, and T.~Lillicrap,
  ``Meta-learning with memory-augmented neural networks,'' in
  \emph{International Conference on Machine Learning}, 2016, pp. 1842--1850.

\bibitem{munkhdalai2017meta}
T.~Munkhdalai and H.~Yu, ``Meta networks,'' in \emph{International Conference
  on Machine Learning}, 2017, pp. 2554--2563.

\bibitem{nichol2018first}
A.~Nichol, J.~Achiam, and J.~Schulman, ``On first-order meta-learning
  algorithms,'' \emph{arXiv preprint arXiv:1803.02999}, 2018.

\bibitem{rajeswaran2019meta}
A.~Rajeswaran, C.~Finn, S.~M. Kakade, and S.~Levine, ``Meta-learning with
  implicit gradients,'' in \emph{Advances in Neural Information Processing
  Systems}, 2019, pp. 113--124.

\bibitem{khodadadeh2019unsupervised}
S.~Khodadadeh, L.~Boloni, and M.~Shah, ``Unsupervised meta-learning for
  few-shot image classification,'' in \emph{Advances in Neural Information
  Processing Systems}, 2019, pp. 10\,132--10\,142.

\bibitem{flennerhag2019meta}
S.~Flennerhag, A.~A. Rusu, R.~Pascanu, H.~Yin, and R.~Hadsell, ``Meta-learning
  with warped gradient descent,'' \emph{arXiv preprint arXiv:1909.00025}, 2019.

\bibitem{kingma2014adam}
D.~P. Kingma and J.~Ba, ``Adam: A method for stochastic optimization,''
  \emph{arXiv preprint arXiv:1412.6980}, 2014.

\bibitem{bertinetto2018meta}
L.~Bertinetto, J.~F. Henriques, P.~H. Torr, and A.~Vedaldi, ``Meta-learning
  with differentiable closed-form solvers,'' \emph{arXiv preprint
  arXiv:1805.08136}, 2018.

\bibitem{ren2018meta}
M.~Ren, E.~Triantafillou, S.~Ravi, J.~Snell, K.~Swersky, J.~B. Tenenbaum,
  H.~Larochelle, and R.~S. Zemel, ``Meta-learning for semi-supervised few-shot
  classification,'' \emph{arXiv preprint arXiv:1803.00676}, 2018.

\bibitem{mtnet}
Y.~Lee and S.~Choi, ``Gradient-based meta-learning with learned layerwise
  metric and subspace,'' in \emph{International Conference on Machine
  Learning}, 2018, pp. 2927--2936.

\bibitem{finn2018metalearning}
C.~Finn and S.~Levine, ``Meta-learning and universality: Deep representations
  and gradient descent can approximate any learning algorithm,'' in
  \emph{International Conference on Learning Representations}, 2018.

\bibitem{Hierarchically}
H.~Yao, Y.~Wei, J.~Huang, and Z.~Li, ``Hierarchically structured
  meta-learning,'' in \emph{International Conference on Machine Learning},
  2019, pp. 7045--7054.

\end{thebibliography}

\begin{IEEEbiographynophoto}{Jathushan Rajasegaran} is currently pursuing a PhD at University of California, Berkeley. Previously, he worked at the Inception Institute of Artificial Intelligence (IIAI), UAE as a Research Associate. He completed his BS from University of Moratuwa, Srilanka and was awarded Gold Medal for his excellent performance. He also received the first prize in IEEE Myron Zucker Design Contest 2017. He has published his research in top venues such as CVPR, WWW and NeurIPS. His research interest lies in computer vision.

\end{IEEEbiographynophoto}\vspace{-2em}

\begin{IEEEbiographynophoto}{Salman Khan}
 received the Ph.D. degree from The University of Western Australia, in 2016. His Ph.D. thesis received an honorable mention on the Deans List Award. From 2016 to 2018, he was a Research Scientist with Data61, CSIRO. He has been a Senior Scientist with Inception Institute of Artificial Intelligence, since 2018, and an Adjunct Lecturer with Australian National University, since 2016. He has served as a program committee member for several premier conferences, including CVPR, ICCV, and ECCV. In 2019, he was awarded the outstanding reviewer award at CVPR and the best paper award at ICPRAM 2020. His research interests include computer vision and machine learning.
\end{IEEEbiographynophoto}\vspace{-2em}

\begin{IEEEbiographynophoto}{Munawar Hayat}
 received his PhD from The University of Western Australia (UWA). His PhD thesis received multiple awards, including the Deans List Honorable Mention Award and the Robert Street Prize. After his PhD, he joined IBM Research as a postdoc and then moved to the University of Canberra as an Assistant Professor. He is currently a Senior Scientist at Inception Institute of Artificial Intelligence, UAE. Munawar was granted two US patents, and has published over 30 papers at leading venues in his field, including TPAMI, IJCV, CVPR, ECCV and ICCV. His research interests are in computer vision and machine/deep learning.
\end{IEEEbiographynophoto}\vspace{-2em}

\vfill

\begin{IEEEbiographynophoto}{Fahad Shahbaz Khan}
is currently a Lead Scientist at the Inception Institute of Artificial Intelligence (IIAI), Abu Dhabi, United Arab Emirates and an Associate Professor (Universitetslektor + Docent) at Computer Vision Laboratory, Linkoping University, Sweden. He received the M.Sc. degree in Intelligent Systems Design from Chalmers University of Technology, Sweden and a Ph.D. degree in Computer Vision from Computer Vision Center Barcelona and Autonomous University of Barcelona, Spain. He has achieved top ranks on various international challenges (Visual Object Tracking VOT: 1st 2014 and 2018, 2nd 2015, 1st 2016; VOT-TIR: 1st 2015 and 2016; OpenCV Tracking: 1st 2015; 1st PASCAL VOC Segmentation and Action Recognition tasks 2010). He received the best paper award in the computer vision track at IEEE ICPR 2016.  His research interests include a wide range of topics within computer vision and machine learning. He serves as a regular program committee member for leading computer vision and artificial intelligence conferences such as CVPR, ICCV, and ECCV.
\end{IEEEbiographynophoto}\vspace{-2em}

\begin{IEEEbiographynophoto}{Mubarak Shah}
is the UCF Trustee chair professor and the founding director of the Center for Research in Computer Vision at the University of Central Florida (UCF). He is a fellow of the NAI, IEEE, AAAS, IAPR, and SPIE. He is an editor of an international book series on video computing, was editor-in-chief of Machine Vision and Applications journal, and an associate editor of ACM Computing Surveys journal. He was the program cochair of CVPR 2008, an associate editor of the IEEE T-PAMI, and a guest editor of the special issue of the International Journal of Computer Vision on Video Computing. His research interests include video surveillance, visual tracking, human activity recognition, visual analysis of crowded scenes, video registration, UAV video analysis, and so on. He has served as an ACM distinguished speaker and IEEE distinguished visitor speaker. He is a recipient of ACM SIGMM Technical Achievement award; IEEE Outstanding Engineering Educator Award; Harris Corporation Engineering Achievement Award; an honorable mention for the ICCV 2005 Where Am I? Challenge Problem; 2013 NGA Best Research Poster Presentation; 2nd place in Grand Challenge at the ACM Multimedia 2013 conference; and runner up for the best paper award in ACM Multimedia Conference in 2005 and 2010. At UCF he has received Pegasus Professor Award; University Distinguished Research Award; Faculty Excellence in Mentoring Doctoral Students; Scholarship of Teaching and Learning award; Teaching Incentive Program award; Research Incentive Award.
\end{IEEEbiographynophoto}\vspace{-2em}
\vfill

%






\end{document}